# TENSOR-BASED MULTI-MODALITY FEATURE SELECTION AND REGRESSION FOR ALZHEIMER'S DISEASE DIAGNOSIS

Jun Yu[1], Zhaoming Kong[1], Liang Zhan[2], Li Shen[3], and Lifang He[1]

[1]Department of Computer Science and Engineering, Lehigh University, Bethlehem, Pennsylvania, USA
juy220@lehigh.edu, zhk219@lehigh.edu, and lih319@lehigh.edu

[2]Department of Electrical and Computer Engineering, University of Pittsburgh, Pittsburgh, Pennsylvania, USA
liang.zhan@pitt.edu

[3]Department of Biostatistics, Epidemiology and Informatics, The Perelman School of Medicine, University of Pennsylvania, Philadelphia, Pennsylvania, USA
li.shen@pennmedicine.upenn.edu

## ABSTRACT

*The assessment of Alzheimer's Disease (AD) and Mild Cognitive Impairment (MCI) associated with brain changes remains a challenging task. Recent studies have demonstrated that combination of multi-modality imaging techniques can better reflect pathological characteristics and contribute to more accurate diagnosis of AD and MCI. In this paper, we propose a novel tensor-based multi-modality feature selection and regression method for diagnosis and biomarker identification of AD and MCI from normal controls. Specifically, we leverage the tensor structure to exploit high-level correlation information inherent in the multi-modality data, and investigate tensor-level sparsity in the multilinear regression model. We present the practical advantages of our method for the analysis of ADNI data using three imaging modalities (VBM-MRI, FDG-PET and AV45-PET) with clinical parameters of disease severity and cognitive scores. The experimental results demonstrate the superior performance of our proposed method against the state-of-the-art for the disease diagnosis and the identification of disease-specific regions and modality-related differences. The code for this work is publicly available at https://github.com/junfish/BIOS22.*

## 1. INTRODUCTION

Alzheimer's Disease (AD) is one of the most common and incurable neurodegenerative diseases [1], which can result in progressive cognitive decline and behavioral impairment, and even cause death in severe cases. A recent report shows that 26.6 million AD patients exist in the world and 1 out of 85 people will be living with AD by 2050 [2]. Thus, for timely therapy that might be effective to slow the disease progression, it is important for early diagnosis of AD and its prodromal stages like Mild Cognitive Impairment (MCI). In particular, MCI is attractive because it represents a transitional stage between normal aging and dementia [3]. Evidence shows that patients with MCI have a high risk to convert into AD, about 12% per year, while healthy controls convert at a 1–2% rate [4].

Recent studies have showed great promise for the use of multi-modality imaging in predicting the progression of AD markers and clinical diagnosis [5–7], such as Magnetic Resonance Imaging (MRI) and Positron Emission Tomography (PET). The results indicate that multi-modality imaging data contain complementary information of clinical importance and thus have the potential to improve the performance of AD diagnosis tasks and provide further insight into the pathophysiology of AD [8]. However, multi-modality imaging data contain noisy and heterogeneous information, making it challenging to effectively incorporate them into quantitative models. It is also challenging to analyze all voxels in the whole brain, as the number of whole-brain voxels usually far exceeds the number of observations available in practice, leading to overfitting.

Recently, various machine learning methods have been proposed for the analysis of multi-modality images in AD studies [6, 9–13]. Feature selection, which searches for a subset of significant features from the original set of features, is critical for effective machine learning, as spurious features can harm the learning process, especially in the presence of multi-modality and high dimensionality data. Furthermore, unlike feature extraction techniques such as principal component analysis (PCA) [14], which project raw data onto a new low-dimensional space, feature selection tends to preserve the interpretability of raw data, as the selected features have clear connections to the original ones. This can help experts to understand which features are relevant to certain disease states.

Feature selection methods are either classifier dependent or classifier independent. The classifier dependent method can potentially provide better performance as it directly makes use of the interaction between features and accuracy [15]. In this respect, several approaches at different levels of complexity have been developed and applied for multi-modality AD diagnosis and risk factor analysis (e.g., [6, 7, 10]). However, existing methods mainly adopt vector space models for multi-modality fusion, which may lead to the loss of valuable information due to vector quantization, and thus degrade the prediction performance.

In this paper, we propose a novel tensor-based method to model the inherent relationship between the features of multi-modality data and perform joint feature selection and tensor regression for AD diagnosis, in which tensor-structured sparsity and low-rankness properties are jointly exploited to learn the interpretable coefficients. We test the feasibility of this method on a large neuroimaging data set from the ADNI cohort using disease severity score and cognitive scores of MMSE and ADAS-13 across three different imaging modalities, including VBM-MRI, FDG-PET, and AV45-PET. Extensive experimental results demonstrate that the proposed method has better prediction performance compared to the vector space models, as well as good interpretability for diagnostic results from selected discriminative regions and modality-related differences.

## 2. MATERIALS AND METHODS

### 2.1. Data Acquisition and Preprocessing

In this work, a total of 692 non-Hispanic Caucasian participants in the Alzheimer's Disease Neuroimaging Initiative (ADNI) database [16] that passed quality control were used for our analysis, including 163 cognitively normal controls (CN), 73 normal controls with significant memory concern (SMC), 214 patients with early MCI (EMCI), 149 patients with late MCI (LMCI), and 93 AD patients. Each subject has three modalities of imaging data, including structural Magnetic Resonance Imaging (VBM-MRI), 18 F-fluorodeoxyglucose Positron Emission Tomography (FDG-PET) and 18 F-florbetapir PET (AV45-PET).

The multi-modality imaging data were aligned to each participant's same visit. The structural MRI scans were preprocessed with voxel-based morphometry (VBM) using the SPM software [17]. Generally, all scans were aligned to a T1-weighted template image, segmented into gray matter (GM), white matter (WM) and cerebrospinal fluid (CSF) maps, normalized to the standard Montreal Neurological Institute (MNI) space as $2 \times 2 \times 2$ mm3 voxels, and were smoothed with an 8 mm FWHM kernel. The FDG-PET and AV45-PET scans were also registered to the same MNI space by SPM. The MarsBaR toolbox[1] was used to group voxels into 116 ROIs defined by Automated Anatomical Labeling (AAL) [18]. ROI-level measures were calculated by averaging all the voxel-level measures within each ROI.

To assess the level of AD's development, we consider three clinical scores as our prediction responses: (1) Disease Severity Score (DSS), in which we treat different stages of AD progression as an index of disease severity (1-CN, 2-SMC, 3-EMCI, 4-LMCI, and 5-AD), (2) AD Assessment Scale–Cognitive 13-item (ADAS-13) [19], with a total score of 85, improving the responsiveness of classic ADAS-Cog [20] for MCI by considering more candidate tasks related to additional cognitive domains, and (3) Mini-Mental State Examination (MMSE) score [21], a widely used screening test of cognitive function among the elderly, with a maximum score of 30 points. We normalize these three prediction scores to the range of [0, 1] so that they have the same scale, and the higher scores indicate the greater severity of the cognitive impairments.

## 2.2. Tensor Construction

Tensors are higher-order generalizations of matrices to multiple indices that can be used to represent multi-dimensional and multi-relational data. To model the inherent relationship and connectivity between the ROIs from multi-modality data for AD/MCI assessment, we investigate three sizes of tensor representations as follows.

- **$116 \times 3$**. We concatenate 116 feature values from all three imaging modalities along an additional modality dimension in the order of VBM, FDG and AV45. Each modality contains 116 ROI-based features, thus resulting in a 2D tensor of size $116 \times 3$ for each subject.

- **$116 \times 116$**. We construct an ROI-to-ROI connectivity matrix based on the pairwise similarity of ROIs, thus obtaining a 2D tensor of size $116 \times 116$ for each subject. For each ROI, we concatenate features from three modalities into a vector of 3 dimensions, denoted as $r_i, i = 1, 2, \cdots, 116$. Then we use the k-Nearest Neighbor ($k$NN) graph [22] to construct the connectivity matrix via the Gaussian similarity function, i.e., $\boldsymbol{X}_{i,j} = \exp\left(-\left\|r_i - r_j\right\|^2 / \sigma^2\right)$, where $\sigma$ is a user defined parameter specifying width. For simplicity, we set $\sigma = 1$ and consider $k = 1, 2, \cdots, 116$ in our paper. In particular, when $k = 116$, it is a fully connected matrix, as each ROI is connected to other ROIs.

- **$116 \times 116 \times 3$**. We construct an ROI-to-ROI connectivity matrix in each modality using the $k$NN graph as above, and concatenate them along an additional modality dimension to generate a 3D tensor of size $116 \times 116 \times 3$ for each subject.

## 2.3. Joint Feature Selection and Tensor Regression

**Figure 1** provides an overview of our proposed method, which uses tensor data as input features to perform feature selection and regression simultaneously.

**Objective Function**. Assume that we have a dataset containing $N$ subjects, each subject is represented by an $M$th-order tensor $\mathcal{X} \in \mathbb{R}^{I_1 \times \cdots \times I_M}$ and is associated with a regression label $y$ indicating disease status. Similar to linear regression, the tensor regression model can be formulated as follows:

$$y = \langle \mathcal{W}, \mathcal{X} \rangle + \varepsilon, \tag{1}$$

---

[1] https://imaging.mrc-cbu.cam.ac.uk/imaging/MarsBar

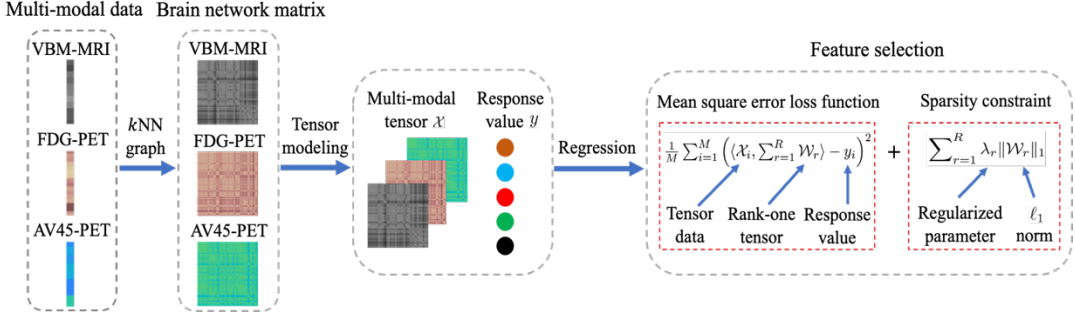

**Figure 1**: The framework of the proposed method. The input data contains three modalities: VBM-MRI, FDG-PET, and AV45-PET, then $k$NN graph is applied to construct tensor data representation. Finally, we regress the response values by using the proposed method.

where $\mathcal{W}$ is the coefficient tensor, $\varepsilon$ is the bias error, and $\langle \cdot, \cdot \rangle$ denotes the inner product operator. To exploit the high-dimensional structure and correlation in the tensor representation, we employ the following sparse and low-rank tensor regression model to solve Eq. (1) inspired by [23].

$$\min_{\mathcal{W}_r} \frac{1}{N} \sum_{i=1}^{N} (\langle \sum_{r=1}^{R} \mathcal{W}_r, \mathcal{X}_i \rangle - y_i)^2 + \sum_{r=1}^{R} \lambda_r \|\mathcal{W}_r\|_1, \quad \text{s.t. CP rank}(\mathcal{W}_r) \leq 1. \quad (2)$$

Where $\mathcal{W}_r = \boldsymbol{w}_r^{(1)} \otimes \cdots \otimes \boldsymbol{w}_r^{(N)}$ is a unit-rank tensor defined upon the CP rank [24], $\otimes$ denotes the outer product operator, and $\lambda_r$ is the regularized parameter that contributes to the sparsity.

Due to the use of a $\ell_1$-norm regularizer in unit-rank tensors, after finding the optimal solution in Eq. (2), we have many zero elements in $\sum_{r=1}^{R} \mathcal{W}_r$, whose corresponding features are not useful in prediction of disease status. Furthermore, by simple arithmetic, we have $\|\mathcal{W}_r\|_1 = \prod_{j=1}^{N} \|\boldsymbol{w}_r^{(j)}\|_1$, i.e., the sparsity of a unit-rank tensor directly leads to the sparsity of its components. This allows us to produce a set of sparse factor components $\boldsymbol{w}_r^{(j)}$ for $j = 1, \cdots, N$ simultaneously. Using this we can examine how each ROI behaves and how each modality contributes to the prediction.

**Optimization**. A common way to solve Eq. (2) is to utilize the alternating least squares (ALS) [25]. However, in this way it is difficult to estimate $R$ and $\lambda_r$. Here we adopt a divide-and-conquer strategy to sequentially solve the following sparse unit-rank estimation problems instead based on the fast stagewise unit-rank tensor factorization (SURF) algorithm [23], which can automatically estimate $\lambda_r$ and is easy to tune the parameter $R$.

$$\min_{\mathcal{W}_r} \frac{1}{N} \sum_{i=1}^{N} (\langle \mathcal{W}_r, \mathcal{X}_i \rangle - y_i^r)^2 + \lambda_r \|\mathcal{W}_r\|_1, \quad \text{s.t. CP rank}(\mathcal{W}_r) \leq 1 \quad (3)$$

where $r$ is the sequential number of the unit-rank terms and $y_i^r$ is the current residue of response with

$$y_i^r := \begin{cases} y_i & \text{if } r = 1 \\ y_i^{r-1} - \langle \mathcal{W}_{r-1}, \mathcal{X}_i \rangle, & \text{otherwise,} \end{cases} \quad (4)$$

Where $\mathcal{W}_{r-1}$ is the estimated unit-rank tensor in the $(r-1)$-th step. The final estimator can be obtained as $\mathcal{W} = \sum_{r=1}^{R} \mathcal{W}_r$.

## 3. RESULTS

## 3.1. Experimental Settings

**Competing Methods.** To show the validity of the proposed method, we compared our method with the PCA feature extraction method followed by linear regression (PCA+LR), and three representative feature selection methods—Lasso [25], Elastic Net (ENet) [26], and Group Lasso (gLasso) [27].

**Evaluation Metrics.** For the quantitative performance evaluation, we employed two metrics: (1) root mean squared prediction error (RMSE), measuring the deviation between the ground truth response and the predicted values, and (2) sparsity of coefficients, defined as the percentage (%) of the number of zero elements to the total number of elements in the coefficients.

**Implementation Details.** For model validation, subjects were randomly split into training and test sets in the ratio of 5:1. The hyperparameters of all methods were optimized using 5-fold cross validation on the training set. Specifically, we traversed all possible percentage of the principal components to realize the best results in PCA+LR. The regularizer parameter of Lasso was selected from $\{0.1, 0.2, \cdots, 1\}$. For the ENet, the weight of lasso ($\ell_1$) versus ridge ($\ell_2$) optimization was learned from $\{0.1, 0.2, \cdots, 1\}$. For the gLasso, the features are grouped by modalities or ROIs, and the parameter-wise and group-wise regularisation penalty are selected via a grid search ranging from $\{10^{-6}, 10^{-5}, \cdots, 10^1\}$, then applying a more fine-grained searching (i.e., $\{0.1, 0.2, \cdots, 1\}$) after the magnitude is determined. In our proposed method, the rank $R$ is incrementally learned from 1 to 70 with step size 1. To avoid randomness from affecting the experiments, five trails were conducted to record mean value and standard deviation of final results for each method.

## 3.2. Experimental Results

**Table 1** shows the performance comparison of five methods. It is clear that the proposed method consistently outperforms the competing methods in terms of both RMSE and Sparsity. Specifically, we observe the following results.

- The prediction performance of our method improves consistently when the feature sizes or dimensions of input data grow, which proves the connectivity information among ROIs in the brain is helpful to AD assessment. The gLasso outperforms the other three baselines and performs better with the higher-order tensor data, because the group variables can consider the structure information to some extent. In contrast, the Lasso and ENet struggle to handle higher dimensional data since vectorization may lead to a certain loss of structural information. These results indicate not only the importance of considering multi-modality information and constructing connectivity among ROIs in the brain data, but also the effectiveness of maintaining the dimensionality in tensor space during the learning process.
- PCA+LR is more consistent to achieve better results with higher dimensionality of input data than lasso-based baselines, bacause the $k$NN graph-based data construction introduces the redundant and noisy information with effective connectivity information and PCA is effective in filtering out unwanted information. However, PCA is a feature extraction method that destroys the structure of original feature space and turns it into principal components, making it difficult to identify significant ROIs of clinical importance. It further validates that our method is robust via seeking a low-rank and sparse tensor space that is helpful to feature selection and noise removal [23].
- Due to the usage of joint sparsity constraint in our objective function, the proposed method can produce the best predictive values and achieve the sparsest solution simultaneously, which increases the interpretability of our model for diagnostic results. More detailed medical significance will be shown in the following qualitative analysis section.

**Table 1**: Performance comparison over different tensor structures on ADNI dataset. The results of mean values and standard deviation (mean ± std) are calculated across 5 trails. ↓ means the lower the better, and ↑ means the higher the better.

| Features | Scores | Metrics | PCA+LR | Lasso | ENet | gLasso | Proposed |
|---|---|---|---|---|---|---|---|
| 116 × 3 | DSS | RMSE↓ | 0.39 ± 0.02 | 0.37 ± 0.01 | 0.36 ± 0.02 | 0.33 ± 0.01 | **0.31 ± 0.01** |
| | | Sparsity↑ | 0.00 ± 0.00 | 85.12 ± 4.14 | 77.41 ± 7.54 | 88.39 ± 0.56 | **97.16 ± 0.43** |
| | ADAS-13 | RMSE↓ | 0.20 ± 0.02 | 0.23 ± 0.01 | 0.23 ± 0.01 | 0.18 ± 0.01 | **0.17 ± 0.00** |
| | | Sparsity↑ | 0.00 ± 0.00 | 85.17 ± 2.76 | 82.82 ± 2.52 | 91.21 ± 0.88 | **98.56 ± 0.41** |
| | MMSE | RMSE↓ | 0.27 ± 0.01 | 0.26 ± 0.20 | 0.26 ± 0.02 | 0.24 ± 0.02 | **0.22 ± 0.02** |
| | | Sparsity↑ | 0.00 ± 0.00 | 87.70 ± 6.05 | 85.23 ± 5.46 | 91.95 ± 1.06 | **96.67 ± 0.16** |
| 116 × 116 | DSS | RMSE↓ | 0.36 ± 0.03 | 0.38 ± 0.01 | 0.38 ± 0.01 | 0.31 ± 0.02 | **0.29 ± 0.01** |
| | | Sparsity↑ | 0.00 ± 0.00 | 98.34 ± 0.55 | 97.39 ± 1.74 | 97.69 ± 0.47 | **99.75 ± 0.04** |
| | ADAS-13 | RMSE↓ | 0.23 ± 0.01 | 0.23 ± 0.01 | 0.23 ± 0.01 | 0.18 ± 0.01 | **0.16 ± 0.01** |
| | | Sparsity↑ | 0.00 ± 0.00 | 98.77 ± 0.09 | 97.99 ± 0.80 | 98.86 ± 0.30 | **99.89 ± 0.02** |
| | MMSE | RMSE↓ | 0.29 ± 0.02 | 0.27 ± 0.01 | 0.27 ± 0.01 | 0.22 ± 0.02 | **0.20 ± 0.02** |
| | | Sparsity↑ | 0.00 ± 0.00 | 99.10 ± 0.24 | 98.38 ± 0.86 | 98.84 ± 0.26 | **99.64 ± 0.03** |
| 116 × 116 × 3 | DSS | RMSE↓ | 0.34 ± 0.02 | 0.37 ± 0.02 | 0.36 ± 0.02 | 0.32 ± 0.01 | **0.28 ± 0.01** |
| | | Sparsity↑ | 0.00 ± 0.00 | 99.38 ± 0.18 | 98.42 ± 0.76 | 98.49 ± 0.25 | **99.97 ± 0.01** |
| | ADAS-13 | RMSE↓ | 0.19 ± 0.02 | 0.23 ± 0.01 | 0.22 ± 0.01 | 0.18 ± 0.01 | **0.14 ± 0.01** |
| | | Sparsity↑ | 0.00 ± 0.00 | 99.55 ± 0.06 | 99.35 ± 0.17 | 98.85 ± 0.25 | **99.97 ± 0.00** |
| | MMSE | RMSE↓ | 0.22 ± 0.02 | 0.26 ± 0.02 | 0.26 ± 0.02 | 0.21 ± 0.02 | **0.18 ± 0.02** |
| | | Sparsity↑ | 0.00 ± 0.00 | 99.77 ± 0.05 | 99.62 ± 0.12 | 98.70 ± 0.44 | **99.97 ± 0.01** |

### 3.3. Explanation Analysis for Feature Selection

**Heatmaps of Discriminative ROIs and Modalities**. To understand the effectiveness of feature selection, we explore the most discriminative regions using features identified by compared methods on the original data of $116 \times 3$. We locate the most discriminative regions based on the selected frequency of each region. **Figure 2** illustrates the average coefficient weights of four feature selection methods across five trials. It is obvious that our method achieves more sparse solutions and pays higher attention to the VBM modality as it is a useful approach for investigating neurostructural brain changes in dementia. The top 10 selected brain regions by our method are: Temporal-Pole-Sup-L, Hippocampus-L, Vermis-10, Hippocampus-R, Temporal-Pole-Mid-L, Angular-L, Cerebelum-10-L, Pallidum-R, Cerebelum-10-R, Caudate-L, Cingulum-Post-L. Most of these selected regions are known to be highly related to AD and MCI in previous studies [28-32]. It is evident that only our method can identify the areas of Hippocampus and Temporal Pole in VBM (as marked in **Figure 2**) which are two critical regions relevant to AD pathology. Furthermore, we also use the factor components learned by our method to measure the contributions of each modality for prediction. According to the average absolute value of $w_r^{(2)}$ for $r = 1, 2, \cdots, 70$, we find that the three modalities are ranked as follows: VBM > AV45 > FDG, which coincide with our results in **Figure 2** and the previous findings [33].

**Colormaps for Physical Brain Regions**. We use the BrainNet Viewer [34] to visualize the brain structure and highlight the regions that the proposed method used to make the predictions against other compared methods, see **Figure 3**. It can be seen that our method is more sparse and uses more relevant ROIs to assess the progression of AD (as marked in **Figure 3** (d)).

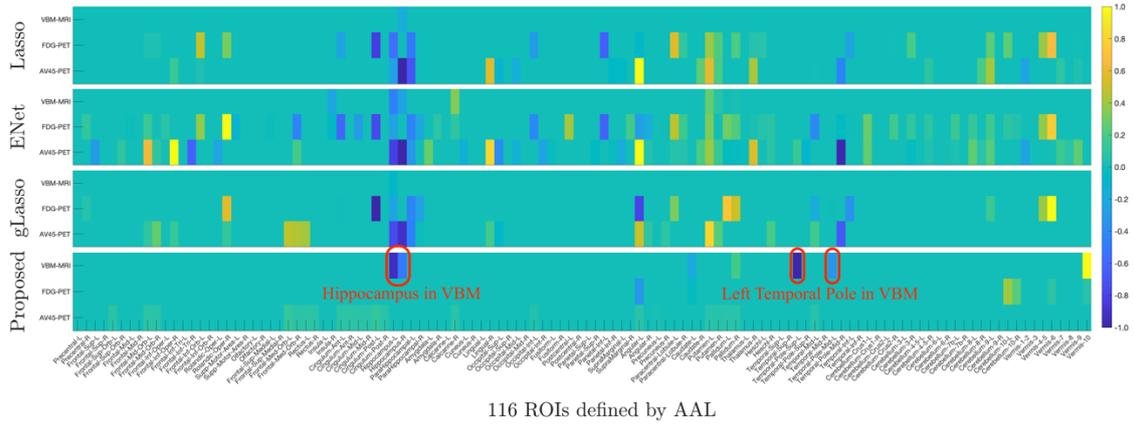

**Figure 2**: Comparison of coefficient weights in terms of each imaging modality across five trials. Each row corresponds to a feature selection method: Lasso, ENet, gLasso, and our proposed method (from top to bottom). Within each panel, there are three rows corresponding to three imaging modalities, i.e., VBM, FDG, and AV45.

### 3.4. Hyperparameter Analysis

We further analyse the influence of two important hyperparameters in our method: $k$ and $R$, where $k$ controls the neighborhood information in the $k$NN graph, and $R$ controls the number of rank-one tensors that are required to approximate the tensor coefficient. We firstly vary the number of neighbors $k$ in $k$NN graph data construction to explore the robustness of our method to the data variation. As shown in **Figure 4** (a), it can be noticed that 1) the higher dimensional data (red line) always outperforms low dimensional data (blue line) across nearly all of possible $k$ values, and 2) the best results are consistently achieved in fully connected graphs (i.e., $k = 116$). The results indicate that our method can make full use of high-order relations among ROIs and modalities. Furthermore, we traverse CP-rank $R$ via a step size 1 to investigate its influence on our model performance, as it plays an important role for feature selection. As shown in **Figure 4** (b), we can observe that 1) the proposed model consistently performs better with the increase of $R$ and tends to be stable with ∼60, and 2) the sparsity of coefficient is reduced when $R$ is increased. The reasons behind these results are because a higher value of $R$ implies that more non-zero values of coefficients are included, and the most relevant features are selected in the beginning itself and the later attributes do not contribute much to the prediction performance. This shows that users of our model can adaptively make a balance between the accuracy and sparsity by controlling $R$.

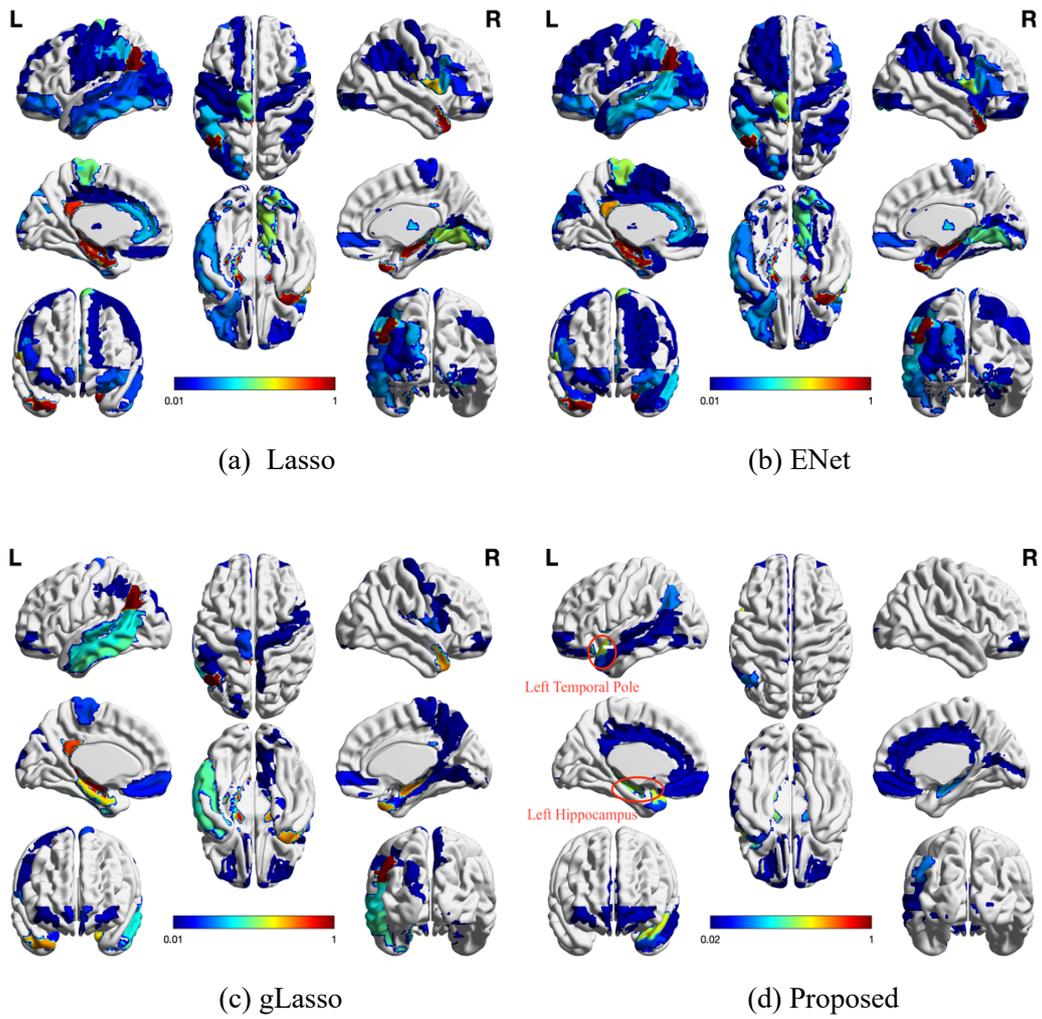

**Figure 3**: The colormaps of 116 ROIs on the physical brains to the corresponding sparse solutions of each feature selection method. Each method shows the full eight-brain views, in which the first row from left to right are lateral view of left hemisphere, topside, lateral view of right hemisphere, the second row from left to right are medial view of left hemisphere, bottom side, medial view of right hemisphere, and the third row are frontal side and backside.

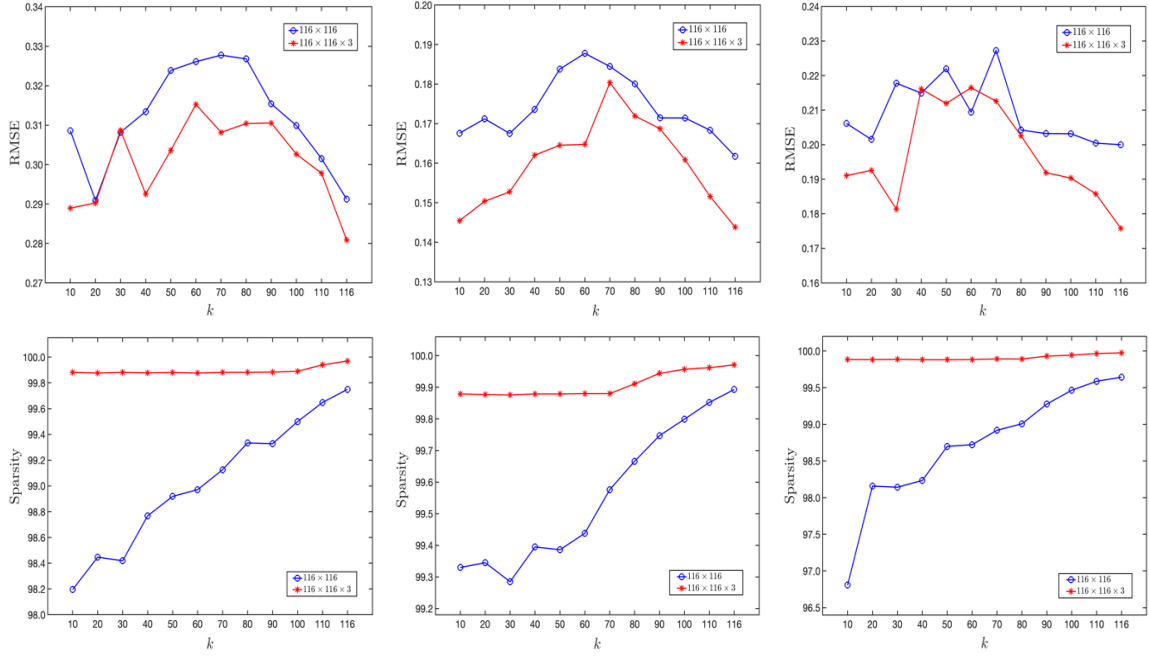

(a) Influence of $k$ on RMSE and Sparsity of DSS, ADAS-13 and MMSE ($R = 60$).

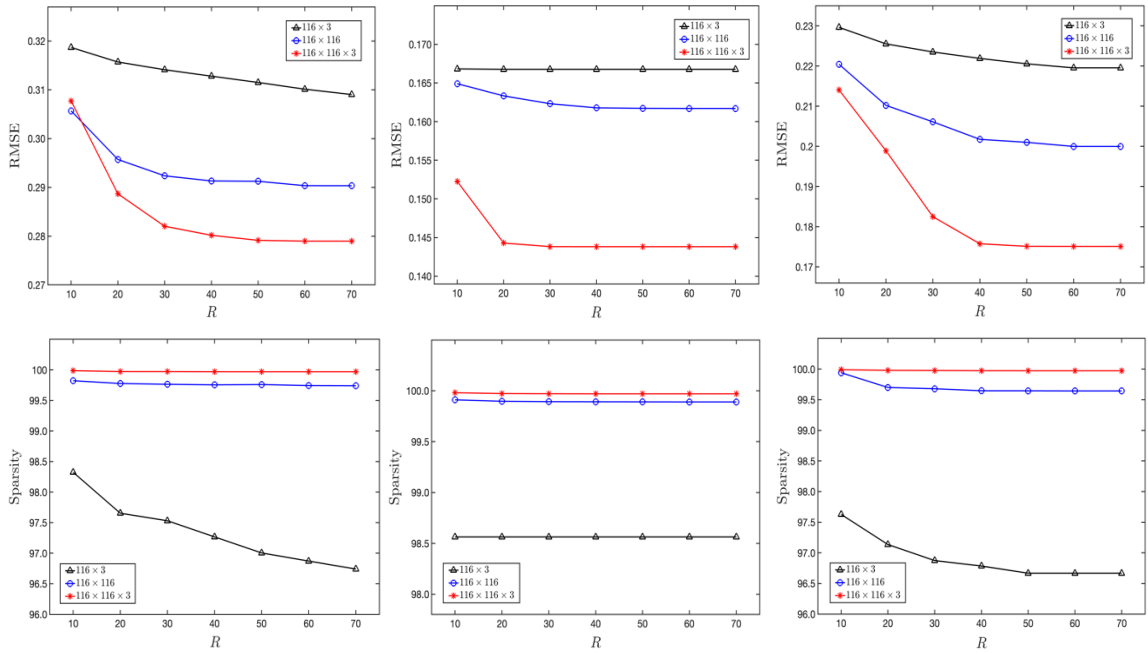

(b) Influence of $R$ on RMSE and Sparsity of DSS, ADAS-13 and MMSE ($k = 116$).

**Figure 4**: Influence of different hyperparameters on model performance associated to DSS, ADAS-13, and MMSE (from left to right). (a) The $k$ value in $k$NN graph to construct the brain network matrix; (b) the tensor CP-rank $R$ to control the low-rank and sparse property of coefficient weights.

## 4. CONCLUSIONS

In this paper, we proposed a joint feature selection and tensor regression model for the prediction of AD-related clinical scores and corresponding biomarker identification, in which tensor-structured sparsity and low-rankness properties are simultaneously exploited to learn the interpretable coefficients. We investigated three different tensor representations to model multi-modality imaging data based on the ROI-level measures with the ADNI database. Our extensive experimental results validated that the proposed method can successfully identify biomarkers related to AD and achieve higher predictive performance than traditional sparse regression methods. Our approach is of wide general interest as it can be applied to other diseases when multi-modality data are available. In the future work, we will extend this approach to simultaneous multiple regression analysis for jointly modelling multiple responses and identifying the importance of regions with multiple clinical scores at the same time.

## ACKNOWLEDGEMENTS

This work is supported in part by NIH grants R01AG071243, R01MH125928, R01LM013463, R01AG066833 and U01AG068057, NSF grants IIS 2045848 and IIS 1837956, ONR grant N00014-18-1-2009 and Lehigh's accelerator grant S00010293.

## Authors

**Jun Yu** is a third-year Ph.D. student in the Department of Computer Science and Engineering at Lehigh University, PA, USA. He received his Bachelor's degree and Master's degree from Shandong University in Software Engineering in 2017 and in Computer Science and Technology in 2020, respectively. His research interests primarily focus on tensor techniques, multi-task learning, and deep learning methods for multi-dimensional data analysis and image computing.

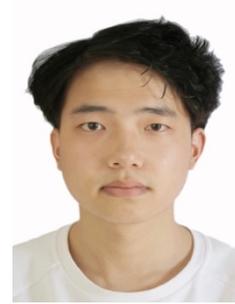

**Zhaoming Kong** is a fourth-year Ph.D. student in the Department of Computer Science and Engineering at Lehigh University, PA, USA. He received his Bachelor's degree and Master's degree from South China University of Technology in Applied Mathematics in 2016 and in Software Engineering in 2019, respectively. His research interests primarily focus on tensor analysis, image processing and deep learning for multidimensional data analysis.

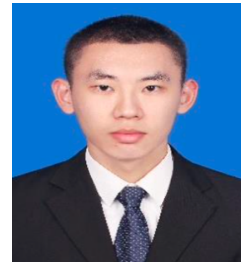

**Liang Zhan** is currently an Assistant Professor in the Departments of Electrical and Computer Engineering and Bioengineering at the University of Pittsburgh, PA, USA. He received his Ph.D. degree in Biomedical Engineering from University of California, Los Angeles (UCLA). His research interests include computational neuroimaging, brain connectomics, machine learning and bioinformatics.

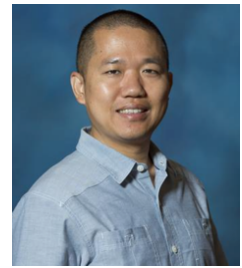

**Li Shen** is a Professor of Informatics in Biostatistics and Epidemiology. His research interests include medical image computing, bioinformatics, machine learning, network science, visual analytics, and big data science in biomedicine. He received his BS degree (Computer Science) at Xi'an Jiao Tong University in 1993, MS degree (Computer Science) at Shanghai Jiao Tong University in 1996, and PhD degree (Computer Science) at Dartmouth College in 2004.

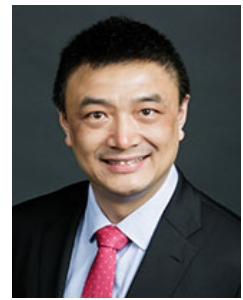

**Lifang He** is currently an Assistant Professor in the Department of Computer Science and Engineering at Lehigh University, PA, USA. She received her Ph.D. degree in Computer Science from South China University of Technology in 2014. Her research interests primarily focus on machine learning, data mining, tensor analysis, with major applications in biomedical data and neuroscience. She has designed several tensor powerful processing and advanced machine learning methods for a range of real-world problems. Her work have been featured by major publications such as WIREs Computational Molecular Science, Neural Networks, TKDE, TIP, NeurIPS, and ICML.

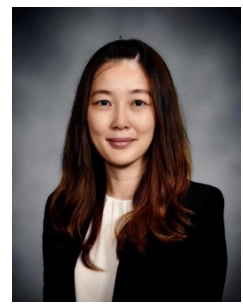